\documentclass[conference]{IEEEtran}
\IEEEoverridecommandlockouts
\usepackage{cite}
\usepackage{amsmath,amssymb,amsfonts}
\usepackage{algpseudocode}
\usepackage{algorithm}
\usepackage{graphicx}
\usepackage{textcomp}
\usepackage{xcolor}

\begin{document}

\title{MPI Planar Correction of Pulse Based ToF Cameras\\
\thanks{The authors are thankful for the support of Analog Devices GMBH Romania, for the equipment list and NVidia for the DGX grade server offered as support to this work. This work was financially supported by the Romanian National Authority for Scientific Research, CNCS-UEFISCDI, project number PN-III-P2-2.1-PTE-2019-0367 and PN-III-P3-3.6-H2020-2020-0060 and by H2020 SeaClear project from the European Union's Horizon 2020 research and innovation programme under grant agreement No 871295.}
}

\author{\IEEEauthorblockN{ Marian-Leontin Pop}
\IEEEauthorblockA{\textit{Automation Department} \\
\textit{Technical University of Cluj-Napoca}\\
Romania \\
popmarianleontin@gmail.com}
\and
\IEEEauthorblockN{ Levente Tamas}
\IEEEauthorblockA{\textit{Automation Department} \\
\textit{Technical University of Cluj-Napoca}\\
Romania \\
Levente.Tamas@aut.utcluj.ro}
}

\IEEEoverridecommandlockouts
\IEEEpubid{\makebox[\columnwidth]{978-1-6654-7933-2/22/\$31.00~
\copyright2022
IEEE \hfill} \hspace{\columnsep}\makebox[\columnwidth]{ }} 

\maketitle

\begin{abstract}
Time-of-Flight (ToF) cameras are becoming popular in a wide span of areas ranging from consumer-grade electronic devices to safety-critical industrial robots. This is mainly due to their high frame rate, relative good precision and the lowered costs. Although ToF cameras are in continuous development, especially pulse-based variants, they still face different problems, including spurious noise over the points or multipath inference (MPI). The latter can cause deformed surfaces to manifest themselves on curved surfaces instead of planar ones, making standard spatial data preprocessing, such as plane extraction, difficult. 
In this paper, we focus on the MPI reduction problem using Feature Pyramid Networks (FPN) which allow the mitigation of this type of artifact for pulse-based ToF cameras. With our end-to-end network, we managed to attenuate the MPI effect on planar surfaces using a learning-based method on real ToF data.
Both the custom dataset used for our model training as well as the code is available on the author's Github homepage.
\end{abstract}


\section{Introduction}

\begin{figure}
    \begin{center}
        \includegraphics[width=8.4cm]{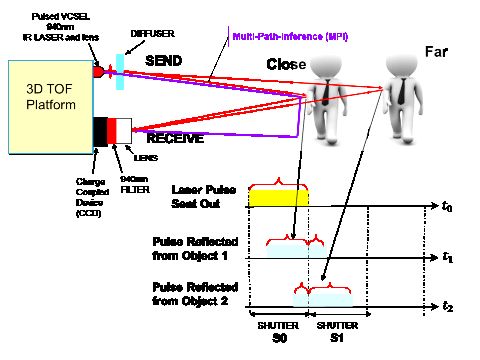}    
        \caption{ToF imaging principle and the MPI effect} 
        \label{fig:tofmpi}
    \end{center}
\end{figure}

Due to the increased popularity of the Time-of-Flight (ToF) cameras in the past decade, the adaption of these devices got as a de-facto standard for the spatial perception in the robotics community. This is due to their relative low cost, high frame-rate and good accuracy. Although they perform well in most environments, some perturbing effects can appear, especially for transparent or translucent objects. For the latter ones, the so-called multi-path inference(MPI), i.e. the reflection of the light from multiple sources towards the receiver can affect the quality of the depth estimation. An illustrative example of the effect of MPI on ToF imaging is shown in Figure \ref{fig:tofmpi}.

For the ToF cameras two different functional approaches exists: the Amplitude Modulated Continuous-Wave Time-of-Flight (AMCW-ToF) and the Pulse-Based Time-of-Flight (PB-ToF) ones \cite{pbtof2018}. The latter are attractive variants in robotics and outdoor perception \cite{tamas20143d}, as they are less affected by external weather conditions \cite{tamas2014robustness}.

The pulse-based ToF cameras on the other hand suffer from the multi-path interference problem, i.e. the multiple reflection of the emitted ray due to different refractions as it can be seen in Figure \ref{fig:tofmpi}. Although with the classical numerical integration techniques, the effect of these perturbations can be reduced, they still affect the quality of the ToF depth imaging\cite{Freedman_2014}. Recently, with the latest deep learning approaches, end-to-end solutions are proposed for specific situations in \cite{agresti2018deep,Guo_2018,Kilho_Son_2016} or by implementing a confidence estimator that identifies perturbed measurements, through a Normalized Convolution Neural Network \cite{Eldesokey_2020} for MPI reduction. Some approaches focus on synthetic data \cite{agresti2019unsupervised} which used a Generative Adversarial Network, but most of these require a huge amount of training data. That is why in many works from the main literature, such as \cite{Guo_2018}, \cite{marco2017deeptof}, \cite{he2017depth}, \cite{agresti2018deep}, they are either using synthetic images in order to train a neural network, thus achieving good results on synthetic images, but based on these images they lack generalization on real data, or they are collecting ground truth data using other expensive high precision reference sensors as in \cite{he2017depth}, \cite{Kilho_Son_2016}. 

In our work we propose an efficient end-to-end deep learning approach for the MPI reduction on planar surfaces. We focus on these larger segments in a ToF image as the extraction of these planes stands on the bases for a number of preprocessing tasks in robotics perception applications \cite{tamas20143d, blaga2021augmented,frohlich2019absolute,kelenyi20223d}. Our solution makes use of a multistep planar estimation with classical sample consensus-based approach, which is used for the training of a feature-pyramid network (FPN) based architecture. The main intuition behind the adoption of this architecture, is the multi-scale estimation capability required for our planar extraction. 

The structure of the paper contains, in addition to the introduction, the related work, focusing on existing approaches for both classical and learning-based variants. In the next section the details regarding the proposed solution are highlighted, and finally the evaluation of the method is discussed for various datasets.

\section{Related Work}

Although many attempts to handle MPI are described in state of the art, none of them tries to focus explicitly on the planar region correction using Convolutional Neural Networks (CNNs). 
A part of them relies on the compensation of continuous wave based ToF cameras using the basic signal propagation equations \cite{fuchs2010multipath}, \cite{jimenez2014modeling} where a continuous wave modulation ToF camera is used. Other approaches are using neural networks to approximate or remove noise from depth images \cite{Kilho_Son_2016}, \cite{marco2017deeptof}, \cite{he2017depth}, \cite{agresti2018deep}, \cite{Eldesokey_2020}, \cite{agresti2019unsupervised}, \cite{Guo_2018}.

In \cite{fuchs2010multipath}, using the signal propagation equations of continuous-wave cameras, a Multi-Path Interference compensation method was proposed, where for each pixel of the depth image, the MPI effect of all the other pixels was subtracted, thus reducing the point cloud distortion. Although this method had good results, it took 10 minutes to calculate per image.
In \cite{jimenez2014modeling} a similar approach is taken, where the authors use signal propagation equations to describe a radiometric model for an MPI iterative compensation algorithm. An MPI free depth map is estimated and taken through the radiometric model where MPI distortion is applied in order to obtain a similar depth map along with the depth map obtained from the sensor, thus when these two depth maps are similar their solution can be considered as an accurate, MPI free, depth map. This solution also claims to have good results, but it still takes a few minutes to compute an accurate depth map, which is not ideal.

In the approach tackled in \cite{Kilho_Son_2016}, the authors made their own dataset by mounting a ToF camera on a robotic arm and with a high precision light sensor that was able to measure a ground truth for the ToF camera. By doing so, they trained two networks: one that maps the measured depth to the real depth and one that learns to detect the objects' boundaries because the MPI effect can also appear near the edges of the objects. Although this approach was good, the data set was limited in the robotic arm setup.

Another approach such as \cite{marco2017deeptof}, is using a synthetic and real dataset in order to train a Convolutional Autoencoder (CAE). They used the real dataset in order to train the encoder, and the synthetic dataset was used to train the decoder of the network, thus obtaining an auto-encoder capable of encoding noisy real images and decoding them into an unchanged depth map, but without the noise, as learned from the synthetic dataset. 
In \cite{he2017depth} they use classical machine learning tools such as SVM for capturing different ToF imaging disturbances such as external light, object reflectance and color. 
The main idea found in \cite{agresti2018deep}, is that they worked with a multi-frequency ToF camera. For estimating the MPI, they used a CNN made of two main parts: a coarse subnetwork and a fine one. They also used ToF data at 20, 50 and 60 MHz frequencies for training. The coarse network takes as input five different input channels, and the fine network the five channels plus the output of the coarse network.The estimated multipath error is then directly subtracted from the ToF depth map, thus resulting in a depth map without multipath distortion, but with other zero-mean error sources. The resulting map is filtered with a 3x3 median filter (removing depth outliers), and a bilateral filter is applied to reduce noise, while preserving edges. They also created a small real dataset used in combination with other synthetic images.

In \cite{Eldesokey_2020}, they proposed a small network with 670K parameters that performs as good as other approaches with millions of parameters. Their idea is based on using a Normalized Convolution Neural Network which is basically taking from the input image only the pixels that have a certain confidence and reconstructing the image using only the pixels with higher certainty. But for NCNN to work, a mask needs to be applied on the input, and in the classical NCNNs it is a binary input confidence mask that considers as a one all the valid input points and zero otherwise, but this approach can lead to artifacts in the output of the CNN. Their solution is to use another network to estimate the input confidence. The output of this NN is fed to an NCNN that outputs a prediction and a confidence estimation map, and those along with the ground truth are fed to the loss function.


Another approach taken in \cite{Guo_2018} was to, also, create a two module deep neural network to correct all the artifacts at the same time. Their first module is a network that is able to attenuate the artifacts caused by objects in motion and second module is a network which tackles the problem of MPI and phase wrapping perturbations. One of the most important things to mention from their work is that they created the FLAT dataset, a very large synthetic dataset made from 2000 Time-of-Flight measurements which includes all these types of artifacts. They also created a network that simulates their used camera hardware (Kinect 2).

\section{Planar MPI reduction}

\subsection{Main idea}

Due to complexity of this problem, we tackle the MPI problem in planar regions found in depth images. While most of the methods explained in the previous section are trying to solve this problem only in the 2D space of a depth image, these artifacts are most visible in the 3D space, that is why we are making use of the camera intrinsic parameters to transform the depth image into a point cloud, during the training phase of an FPN based model and consider a 3D metric. 


The pseudo-code of the proposed method for training the FPN model is presented below:

\begin{algorithm}
\caption{\textbf{Planar correction algorithm}}\label{alg:cap}
\begin{algorithmic}

\State depth\_images = get\_depth\_from\_ToF\_camera()

\For{depth\_image \textbf{in} depth\_images}
    \State pcd = depth\_2\_pcd(depth\_image)
    \State th = 1.7
    \State angle = 20.0
    \State loose\_inliers, loose\_coeff = Plane(pcd, th, angle)
    \State th = 1.3
    \State angle = 15.0
    \State tight\_inliers, tight\_coeff = Plane(pcd, th, angle)
    \State ideal\_plane = points\_2\_plane(loose\_inliers, tight\_coeff)
    \State pcd -= loose\_inliers
    \State rectified\_pcd = pcd + ideal\_plane
    \State gt\_image = pcd\_2\_depth(rectified\_pcd) 
\EndFor
\For{depth\_image, gt\_image}
\State fpn\_model = training(depth\_image, gt\_image)
\EndFor

\end{algorithmic}
\end{algorithm}



\subsection{Proposed network architecture}

\begin{figure}[h]
    \begin{center}
        \includegraphics[width=8cm]{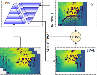}    
        \caption{Proposed FPN based training setup} 
        \label{fig:arch}
    \end{center}
\end{figure}

The input size of this network can be arbitrary (depending on the internal parameters of the ToF camera) and the output results in a proportionally sized feature map with the input image using a fully convolutional approach. The approach is generic in the sense that for the convolutional architecture custom variants can be adopted \cite{molnar2021feature} such as depth or IR images. The construction of these pyramids involves a bottom-up and top-down path with lateral connections \cite{FPN2017}. An intuitive representation for the multiscale architecture for the ToF MPI correction can be seen in Figure \ref{fig:arch}.

\begin{figure}
    \begin{center}
        \includegraphics[width=5.4cm,height=8cm]{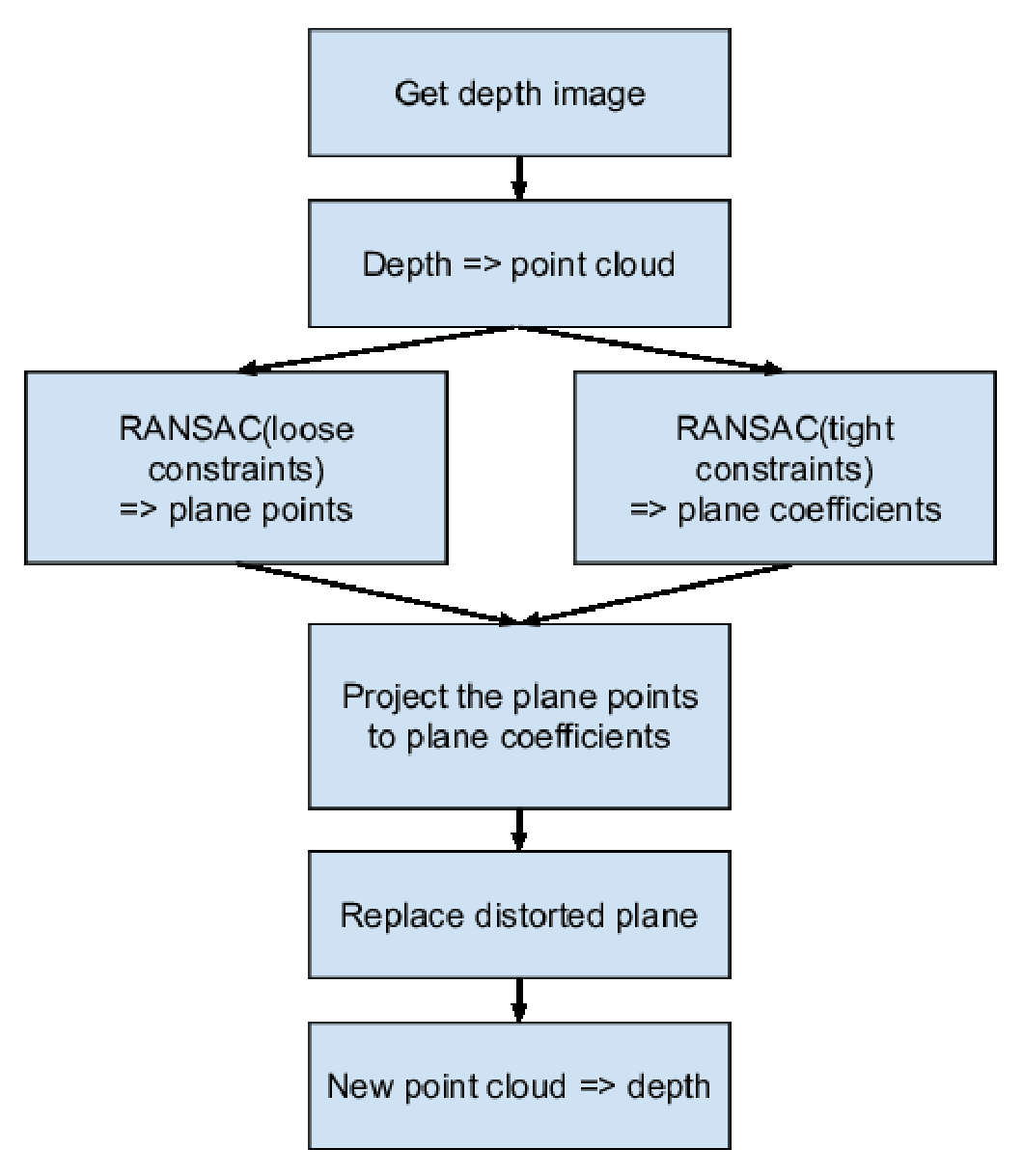}    
        \caption{Network training stages} 
        \label{fig:network_train}
    \end{center}
\end{figure}

As a first step in using this architecture, we load the dataset composed of a set of three-channel images, every channel containing the raw depth image, and the corresponding ground-truth images with the rectified plane. After data loading, we normalize the images in the [0-1] range, to achieve a robust computation in the training process. In the next step, we create a loss function that can minimize the error between the predicted image and the ground truth image. Because our key information about the MPI artifacts is present in the 3D space, we transform the predicted image and the ground truth image in point clouds and only then we compute our loss function, trying to minimize the difference between the two point clouds. A more schematic representation of these steps can be seen in Figure \ref{fig:network_train}. 

\subsection{Dataset generation}

\begin{figure}[h]
    \begin{center}
        \includegraphics[width=8.6cm]{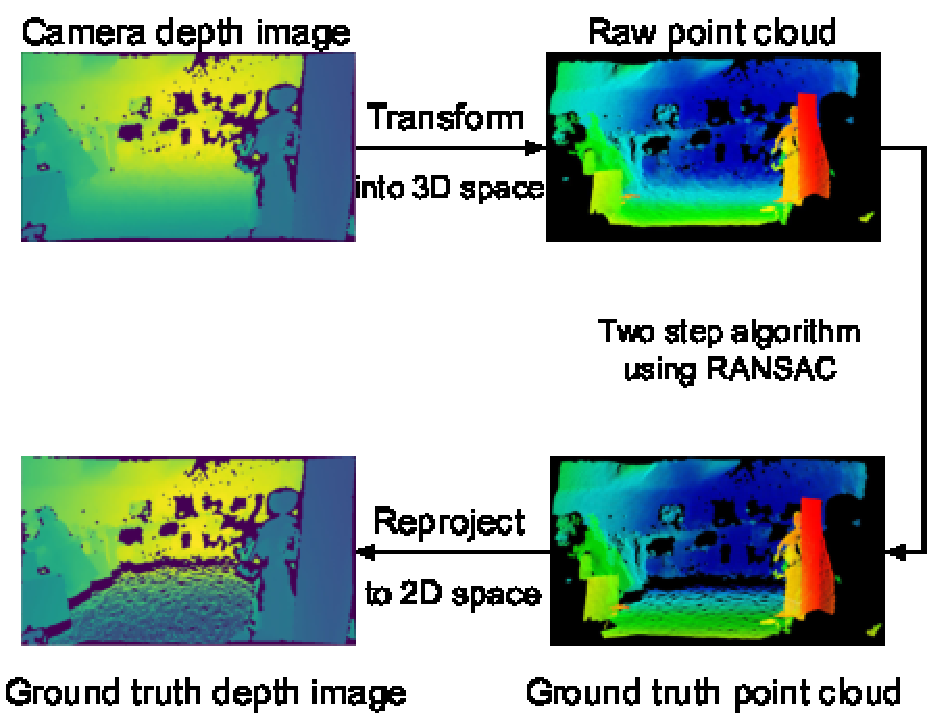}    
        \caption{Ground truth creation flowchart} 
        \label{fig:idea_sketch}
    \end{center}
\end{figure}

In order to implement a neural network which is able to learn to correct these artifacts, we created a dataset with raw depth images provided by the camera, with MPI affected ground floor plane, and the corresponding ground truth for each of these images.

We created the ground truth images by first transforming our 2D images into point clouds, using the camera intrinsic parameters. After this transformation, the artifacts created by the MPI are more relevant in the 3D space. The targeted artifacts consist of groups of points that are not in the same plane, as can be seen in Figure \ref{fig:idea_sketch}. Than we apply a two-stage sample consensus based planar extraction in order to get a close-to-real planar patch estimation.

For the Random Sample Consensus (RANSAC) methods we adopted ready-to-use library solutions from PCL \cite{rusu20113d} and Open3D \cite{zhou2018}.
We constrained the RANSAC segmentation algorithm to identify only those points which belong to the plane that is perpendicular on the $Oy$ axis (i.e. the ground floor plane, in our case). This algorithm has a set of parameters, detailed in pseudocode from section \ref{alg:cap}, which can be adjusted to fit the points in a plane. By adjusting only these parameters of the algorithm, we are not being able to select all the points which belong to the ground floor, due to distorting artifacts. That is why we propose a two step algorithm in order to find the best approximation of the ground floor plane. 

\begin{figure}[h]
    \begin{center}
        \includegraphics[width=8.4cm,height=8cm]{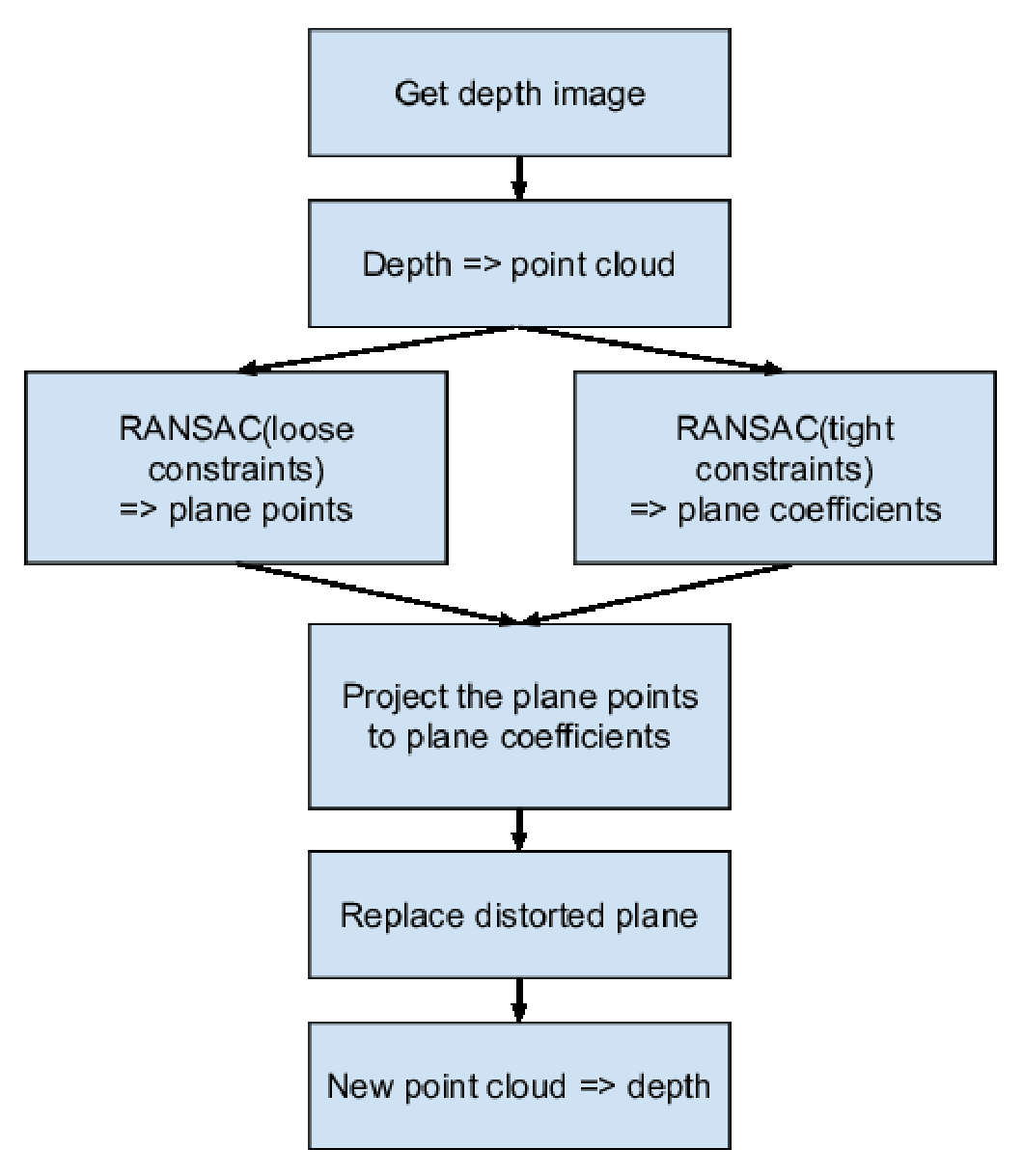}    
        \caption{Steps for creating ground truth data} 
        \label{fig:gt_creation}
    \end{center}
\end{figure}

By doing so, we obtain all the points that belong to the ground floor which we project on the plane with coefficients obtained in the first iteration of the RANSAC segmentation algorithm, thus obtaining a rectified plane in our point cloud. After we obtain this plane, all we need to do is project this point cloud back into the 2D space, in a depth image. A more illustrative representation of this algorithm can be seen in Figure \ref{fig:gt_creation}. 

With the available ground truth planar patches, we can train our model to estimate these artifacts at a run time that is better than the whole algorithm proposed in Figure \ref{fig:gt_creation}, allowing us to have a more efficient MPI planar reduction method.


\subsection{Model training}

Figure \ref{fig:network_train} shows a top view of the required stages of our network training phase. The loss function between the prediction and the ground truth images is computed in the 3D space, thus we use the following equations, empirically determined, in order to minimize the distance between the points and to make full use of the spatial coordinates of the points within the point clouds:

\begin{equation}
    \label{eq:lossX}
    lossX = \frac{1}{n}\sum_{i=1}^{n}{(x_{P_i} - x_{P_i^*})^2}
\end{equation}

\begin{equation}
    \label{eq:lossY}
    lossY = \frac{1}{n}\sum_{i=1}^{n}{(y_{P_i} - y_{P_i^*})^2}
\end{equation}

\begin{equation}
    \label{eq:lossZ}
    lossZ = \frac{1}{n}\sum_{i=1}^{n}{(z_{P_i} - z_{P_i^*})^2}
\end{equation}

\begin{equation}
    \label{eq:lossRMSE}
    loss_{RMSE} = \sqrt{\frac{1}{n}\sum_{i=1}^{n}{ \left(\ln{|z_{P_i}|} - \ln{|z_{P_i^*}|} \right) ^2}} 
\end{equation}

\noindent
where:
\begin{itemize}
    \item $x_P, y_P, z_P$ are the coordinates of point $P$ of the ground truth point cloud
    \item $x_P^*, y_P^*, z_P^*$ are the coordinates of point $P^*$ of the predicted point cloud
    \item $n$ is the total number of ground truth points.
\end{itemize}
By merging the loss functions defined in equations \ref{eq:lossX}, \ref{eq:lossY}, \ref{eq:lossZ} and \ref{eq:lossRMSE} we can accurately state the final form of our loss function:
\begin{equation}
    \label{eq:loss}
    loss = s \cdot loss_{RMSE} \cdot |3 - e ^ {lossX} - e ^ {lossY} - e ^ {lossZ}|
\end{equation}
where: 
\begin{itemize}
    \item $s$ is a hyperparameter used to speed up the training time, $s$ being $100000$ in our case.
\end{itemize}

Both the training and evaluation are done using the Pytorch library v1.7.1 \cite{NEURIPS2019_9015} running with native CUDA 11 support on a 3080 RTX Nvidia GPU. The rest of the PC hardware is consumer-level devices, while the ToF camera is from the ADI with medium depth settings and f=0 filtering parameters. The details regarding the hyperparameters options can be found on GitHub\footnote{https://github.com/Funderburger/mpi-planar-correction.git}.




\subsection{Model evaluation metrics}

\begin{figure}
    \begin{center}
        \includegraphics[width=8.6cm]{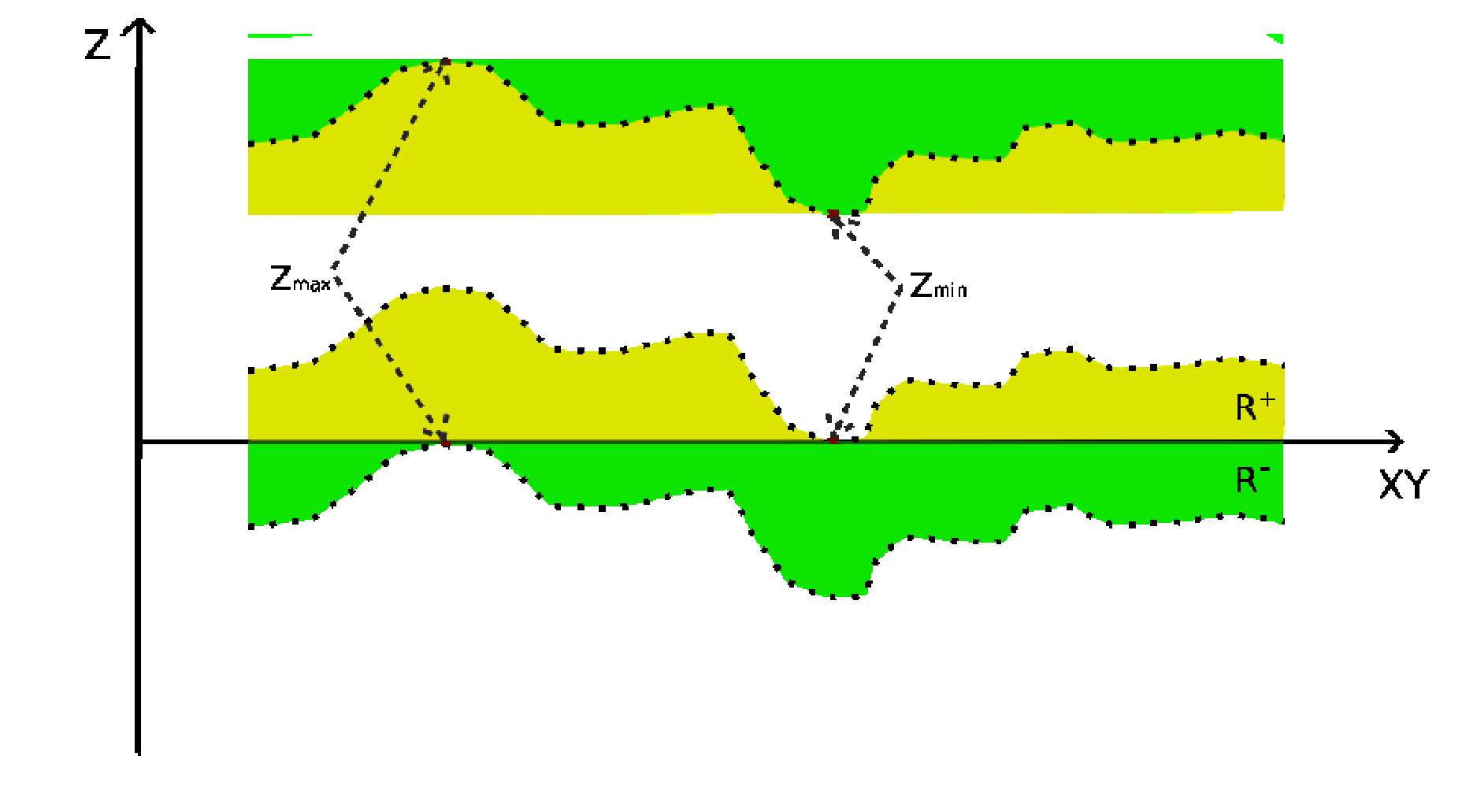}    
        \caption{Visual representation of equation \ref{eq:curv_grad}. The top dotted line represents the (sectioned) plane, parallel with the $XY$ axes. The yellow region ($R^+$) represents the area (volume in 3D) of the plane with respect to XY axes if it is translated with $Z_{min}$ (the middle dotted line) and the green region ($R^-$) represents the area (volume in 3D) of the plane with respect to XY axes if it is translated with $Z_{max}$ (the bottom dotted line)} 
        \label{fig:curv_grad}
    \end{center}
\end{figure}

In order to evaluate our model we created an algorithm for measuring the MPI effect on planar surfaces, i.e planar distortions. First of all, we use the trained model to obtain the estimated pairs for the training set images and the test set images, and from both of these pairs we extract the biggest planar surface found in each of them. Each extracted point cloud, now consisting only of a planar surface, has a set of planar coefficients (determined using the RANSAC algorithm of the Open3D library \cite{zhou2018}) which we use, in order to apply a transformation and a rotation to our plane to bring it in a parallel position with the $XY$ plane. Having the plane in this position, as in Figure \ref{fig:curv_grad}, we compute a global curvature gradient:
\begin{equation}
    \label{eq:curv_grad}
    curv\_grad = \frac{1}{m}\sum_{i=1}^{m}{\frac{|R_i^-|}{|R_i^-| + |R_i^+|}}
\end{equation}

\noindent
where:
\begin{itemize}
    \item $R^-$ represents a vector containing all the $z$ values of the extracted plane below the $XY$ axes;
    \item $R^+$ represents a vector containing all the $z$ values of the extracted plane above the $XY$ axes;
    \item $m$ represents the number of points in the extracted plane.
\end{itemize}

\begin{figure}
    \begin{center}
        \includegraphics[width=8.6cm,height=5cm]{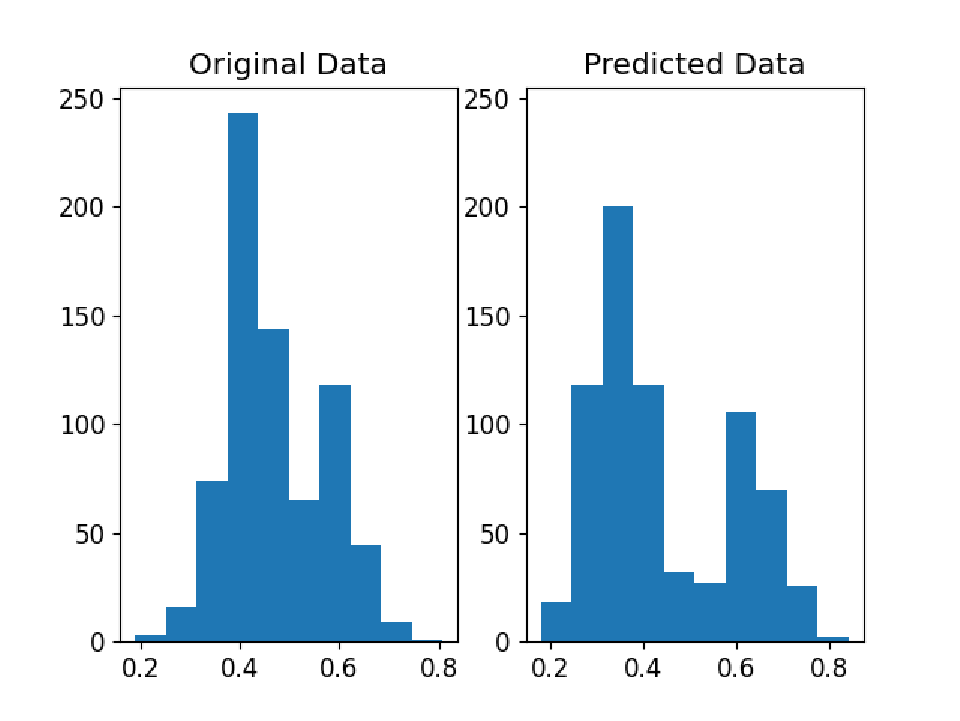}    
        \caption{Training data histograms, with original data std: 0.0978 and predicted and std: 0.1455} 
        \label{fig:super_train_hist}
    \end{center}
\end{figure}

\begin{figure}
    \begin{center}
        \includegraphics[width=8.6cm,height=5cm]{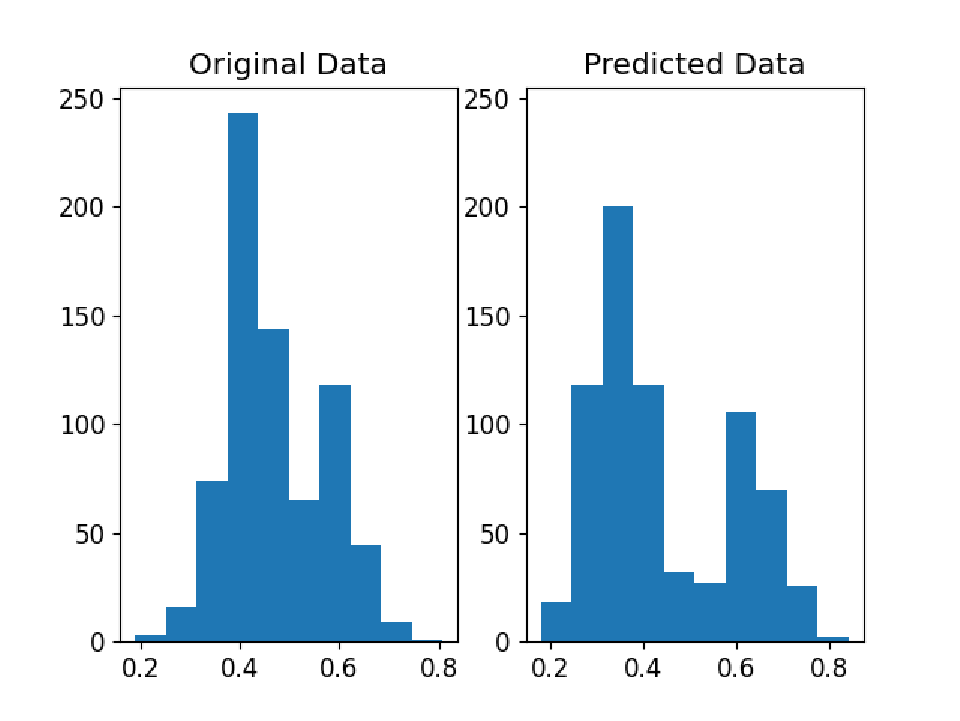}    
        \caption{Test data histograms, with original std: 0.0965 and predicted std: 0.1458} 
        \label{fig:super_test_hist}
    \end{center}
\end{figure}

Using (\ref{eq:curv_grad}) we can now compare the ground floor planes extracted from the raw input images and the ground floor planes extracted from the model predicted images. The gradient will have values in the range [0-1] since it represents (Figure \ref{fig:curv_grad}) what percentage of the whole area, yellow and green, occupies the green area. Thus, by using this metric we can evaluate our model to see how well it performs on the training set and the test set. In Figure \ref{fig:super_train_hist} we created a pair of histograms and their corresponding standard deviation (for raw, respectively predicted data) after we computed the curvature gradient for all the 2128 extracted planes found in the training dataset. The same process was used to create the histograms from Figure \ref{fig:super_test_hist}, but for the 717 planes extracted from the test data set. As can be observed from these two figures, our model decreases the MPI effect on most planar surfaces as it lowers the curvature gradient computed on the predicted data: from the $[0.4 - 0.5]$ range to the $[0.2-0.3]$ range on the training and test datasets, but at the same time, it is not perfect, as can be seen, because it also adds distortion to some planes, thus increasing the number of planes with a curvature gradient in the range $[0.6-0.8]$.



\begin{figure}
    \begin{center}
        \includegraphics[width=8.6cm]{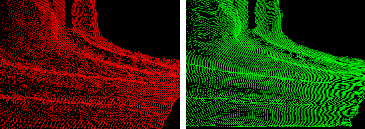}    
        \caption{Indoor, robot case 1: left input point cloud, right predicted point cloud (best visible in color)} 
        \label{fig:c24_case_1}
    \end{center}
\end{figure}

\section{Experimental validation with mobile robot}




\subsection{Indoor setup}
\begin{figure}[h]
    \begin{center}
        \includegraphics[width=8.6cm]{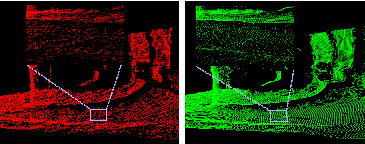}    
        \caption{Indoor, robot case 2: left input point cloud, right predicted point cloud (best visible in color)} 
        \label{fig:c24_case_2}
    \end{center}
\end{figure}
We test our trained model, with the depth camera attached to the P3-AT robot. As can be seen in Figure \ref{fig:c24_case_1}, the model does bring an improvement with regards to correcting the planar surface, even though it is not very visible. A closer look can be seen in Figure \ref{fig:c24_case_2}, where on a small scale, the model removes some of the raw noise given by the camera, which offers a more smooth surface.

\subsection{Outdoor setup}

\begin{figure}[h]
    \begin{center}
        \includegraphics[width=8.6cm]{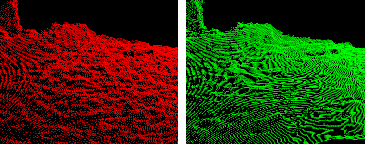}    
        \caption{Outdoor: left input point cloud, right predicted point cloud (best visible in color)} 
        \label{fig:outdoor_case}
    \end{center}
\end{figure}

Despite the fact that we were focusing on the indoor data set, we also performed cross-validation on the outdoor data set too. 
As it can be seen in Figure \ref{fig:outdoor_case}, the data that camera is able to provide is noisy due to direct influence of the sun's light rays, even though the ground is made from solid concrete, which is not a very reflective surface. While our model is trying to rectify the ground, it contributes less to the surface undistortion, which we interpret as the lack of the number of outdoor training images in the dataset and the influence of the strong lightning condition on the scene, affecting the data capture with the ToF camera. 

\section{Conclusion}
In this work we proposed a novel end-to-end deep learning based multi-path-interference reduction for pulse based Time of Flight cameras based on Feature Pyramid Networks. To generate training data, we used a two-stage sampling consensus-based planar patch extraction algorithm. For the loss, we considered various types of empirically determined metrics, while as for the evaluation metric, we constructed our own curvature measuring gradient in order to compare two different planes without being affected by their scales.

As future work, we would like to extend this method towards regular geometric shapes which can be extracted with sampling consensus based methods for reducing the MPI on these surfaces as well.

\bibliographystyle{plain}
\bibliography{ifacconf} 

\end{document}